\documentclass[10pt,twocolumn,letterpaper]{article}
\usepackage{iccv}
\usepackage{times}
\usepackage{epsfig}
\usepackage{graphicx}
\usepackage{amsmath}
\usepackage{amssymb}
\usepackage{bbm}
\usepackage{subcaption}
\usepackage[accsupp]{axessibility}

\usepackage[pagebackref=true,breaklinks=true,letterpaper=true,colorlinks,bookmarks=false]{hyperref}

\iccvfinalcopy 

\def\iccvPaperID{9109} 

\ificcvfinal\fi
\makeatletter
\newcommand{\printfnsymbol}[1]{%
  \textsuperscript{\@fnsymbol{#1}}%
}
\makeatother
\begin{document}


\title{Class Semantics-based Attention for Action Detection}

\author{Deepak Sridhar$^1$\thanks{denotes equal contribution} \thanks{Corresponding author}  \and Niamul Quader$^1$\footnotemark[1] \and Srikanth Muralidharan$^1$\and Yaoxin Li$^{1,2}$\and Peng Dai$^1$\and Juwei Lu$^1$ \\ Huawei Noah's Ark Lab, Canada$^1$ \ \ University of Waterloo$^2$ \\ \tt\footnotesize \{deepak.sridhar1,niamul.quader1,srikanth.muralidharan,yaoxin.li,peng.dai,juwei.lu\}@huawei.com}

%

\maketitle
\ificcvfinal\thispagestyle{empty}\fi

\begin{abstract}
       Action localization networks are often structured as a feature encoder sub-network and a localization sub-network, where the feature encoder learns to transform an input video to features that are useful for the localization sub-network to generate reliable action proposals. While some of the encoded features may be more useful for generating action proposals, prior action localization approaches do not include any attention mechanism that enables the localization sub-network to attend more to the more important features. In this paper, we propose a novel attention mechanism, the Class Semantics-based Attention (CSA), that learns from the temporal distribution of semantics of action classes present in an input video to find the importance scores of the encoded features, which are used to provide attention to the more useful encoded features. We demonstrate on two popular action detection datasets that incorporating our novel attention mechanism provides considerable performance gains on competitive action detection models (e.g., around 6.2\% improvement over BMN action detection baseline to obtain 47.5\% mAP on the THUMOS-14 dataset), and a new state-of-the-art of 36.25\% mAP on the ActivityNet v1.3 dataset. Further, the CSA localization model family which includes BMN-CSA, was part of the second-placed submission at the 2021 ActivityNet action localization challenge. Our attention mechanism outperforms prior self-attention modules such as the squeeze-and-excitation in action detection task.~We also observe that our attention mechanism is complementary to such self-attention modules in that performance improvements are seen when both are used together.
\end{abstract}
\section{Introduction}
The creation of digital videos and the need for video understanding has exploded over the last decade due to widespread availability and presence of digital cameras. Two fundamental components of video understanding are identifying the action components that are present in the video \cite{slowfast,TSM,TSN} and localizing these actions across the temporal axis \cite{RAM,RAPNet,GTAN,GTAD,bmn} and also the spatial axis \cite{he2020,spatiotemporal2013}. For video understanding tasks, it is crucial to learn good video representations, learning relevant encoded features that are useful for the video understanding tasks \cite{slowfast, bmn, bsn}. These rich video representations or encoded features can then be utilized to perform various video understanding tasks (e.g. a localization sub-network can use these features for detecting action segments, etc.). This paper focuses on the task of temporal action detection in untrimmed videos, which has several applications such as content-based video searching \cite{jiang2012}, video highlight generation \cite{soccernet} and surveillance \cite{cristani2007}.

With the availability of large-scale action recognition and detection datasets (e.g. Kinetics-400 \cite{I3D}, ActivityNet v1.3 \cite{activitynet}, etc.) and the availability of high performance computing services, deep learning approaches have achieved enormous success in learning video representations or learning encoders that generate features which are useful for video understanding tasks such as temporal action detection \cite{TSM,slowfast,TSN,c2f}. These encoder learning processes can be very comprehensive - e.g. the encoder learning process of BMN \cite{bmn} (a recently proposed competitive approach) first involves training a TSN-ResNet101 action recognition model \cite{TSN} on Kinetics-400 action recognition task \cite{cuhk} followed by fine-tuning the action recognition model on ActivityNet v1.3 \cite{activitynet}, followed by extracting action class-semantic rich features using the action recognition model, and finally using a localization encoder to encode features that are useful for video understanding tasks such as action localization. These encoded features are processed by the localization sub-network for generating action proposals \cite{zhao2017temporal, bsn, bmn}. It is unlikely that all of these encoded features will be equally useful or important for the localization sub-network for a particular video. In fact, prior works on attention in convolutional neural networks (ConvNets) have shown that attending to more important feature channels \cite{squeeze} or locations \cite{cbam} can accelerate training and usually improve network performances. Despite the possibility that the importance of the encoded feature can vary for different videos and the existing motivation from prior works that have demonstrated improved ConvNet performances via attention mechanisms \cite{cbam,squeeze,transformer}, there are no prior action localization networks that have attention mechanisms for attending to the more important encoded features for accelerating training and improving performance.

\begin{figure*}[!ht]
\centering
\includegraphics[keepaspectratio, width=1.\textwidth]{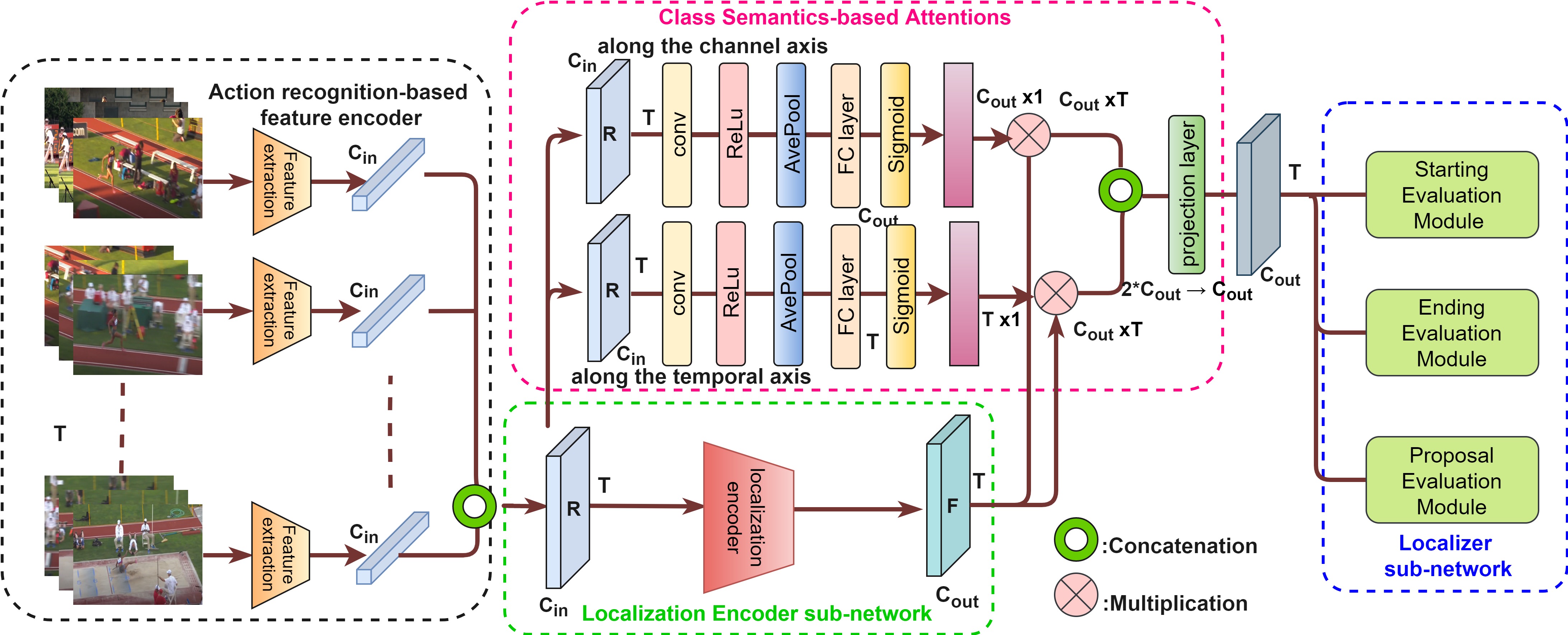}
\caption{The architecture of a generic action detection network with incorporation of our CSA attention mechanism. The generic action detection architecture consists of three major components: (1) an action recognition-based feature encoder that extracts class semantics rich features $R$, (2) a localization encoder sub-network that encodes $R$ to $F$, and (3) a localization sub-network that processes $F$ for generating action proposals. Our attention mechanism learns attention weights from $R$ and applies attention on $F$ both along the channel and the temporal axis, and then fuses the two attention-applied outputs.} 
\label{attn_arch}
\end{figure*}

Incorporating attention at the encoded feature, while not used before for action localization ConvNets, can easily be implemented with prominent self-attention methods (e.g. squeeze-and-excitation (SE) based attention \cite{squeeze}, transformer~\cite{transformer}), which learn the inter-dependencies of the encoded feature to estimate the relative importance of the features. In contrast, we propose a novel attention mechanism that computes the relative importance of the features based on class-specific semantically rich features that are extracted by the action recognition model and that are used at the input of the encoder in the action localization network (Fig.~\ref{attn_arch}). Our rationale on using these class-specific semantically rich features is that the distribution of importance of the encoded feature will likely depend based on which action class (or action classes) the video contains. Evidence for such class-specific dependency of importance distribution was demonstrated in an ablation study in \cite{squeeze} which showed that some specific feature channels were more important than others for one particular class while having low importance for another class within the ImageNet dataset \cite{imagenet}. For action localization tasks, the class-semantics can vary across the temporal axis since different action classes can be present at different time points. Prior works on action localization in the fully and weakly supervised setting (where only class-level supervision is used to learn the temporal boundaries of an action), use class semantic-rich classification features as input and have successfully pushed the state of the art localization performances. This shows that the class-semantics features contain useful information for reliable action localization. To learn the importance of features from the temporally varying class-semantics in videos, our novel attention mechanism jointly learns from both the channel and temporal axes of the encoder input features, and provides attention both along the channel and temporal axes of the encoded features. Our attention mechanism is generic and can be easily applied to prior action localization networks that have an encoder and a localization sub-network (e.g. \cite{bmn,DBG2020arXiv,GTAD}) (Fig.~\ref{attn_arch}). We demonstrate that our attention mechanism considerably improves on baseline ConvNets \cite{bmn,GTAD,DBG2020arXiv} on two major action detection datasets (Thumos\cite{thumos} and ActivityNet v1.3 \cite{activitynet}). Our ablation studies also show that our novel attention mechanism can provide complementary benefits when used together with self-attention mechanisms (such as \cite{squeeze}).

\section{Related Works}
\subsection{Attention in video understanding}

Attention mechanisms are used in a variety of video recognition tasks including fully supervised action recognition~\cite{girdhar2017attentional,long2018attention,gberta_2021_ICML,sharma2015action,weightexcitation}, weakly supervised action detection~\cite{wang2017untrimmednets,nguyen2018weakly,liu2019completeness} and spatio-temporal action localization~\cite{girdhar2019video}. In action recognition models, attention is used for performing different operations, including weighted spatio-temporal pooling or gating~\cite{sharma2015action,girdhar2017attentional, xie2017rethinking,long2018attention}, encoding spatio-temporal features by capturing different kinds of relations and dependencies~\cite{girdhar2019video,gberta_2021_ICML,fan2021multiscale}. In weakly supervised action detection methods~\cite{wang2017untrimmednets,nguyen2018weakly,paul2018w}, attention mechanism is used to rank clips for action detection, with the weights learned from video level classification objectives.  
Quader et al.~\cite{weightexcitation} showed that some feature channels become more important than others during training and introduced an attention mechanism that focused on optimizing those important feature channels to improved action and gesture recognition performances. 
Unlike these methods, our focus in this work is on fully supervised temporal action localization, where we use our attention mechanism to learn from temporal distribution of semantics of action classes in the multi-crop clip representations and apply it on our encoded features that are then processed by a localization sub-network for generating action proposals.

\subsection{Attention in temporal action localization}
The above mentioned attention mechanisms proposed for action recognition tasks~\cite{sharma2015action,wang2017untrimmednets,xie2017rethinking,girdhar2017attentional,weightexcitation} are not directly useful for action localization for the following reasons: (1) action recognition models have spatio-temporal RGB inputs whereas competitive action localization models use encoded temporal inputs, (2) an action recognition model's learnable parameters (including any attention mechanisms inside) are usually not part of the training process of competitive action localization networks due to computational constraints in end-to-end training~\cite{bmn,GTAD,DBG2020arXiv,zhao2020bottom}. In fact, there exists no prior works on applying attention mechanisms in competitive action localization networks. Recent competitive action localization ConvNets are composed of an encoder sub-network that encodes features for downstream localization task and a localization sub-network that generates action proposals. The most straightforward application of attention in action localization networks is perhaps to apply channel-wise self-attention at the encoder output via processing the inter-dependencies of the encoder output using self-attention mechanisms such as SE~\cite{squeeze}, CBAM~\cite{cbam} and transformer~\cite{transformer}. In this paper, we explore such self-attention mechanisms, and more importantly, propose a novel attention mechanism that processes the action class-specific semantically rich temporal features at the input of the localization network to identify channel-wise and temporal-point-wise attention at the encoder output; our rationale - the temporal variation of class-semantics of a video can have important cues on which channels and time-points the localization sub-network should attend to.

\section{Technical Approach}
We first document the layout for a generic action localization architecture (Fig.~\ref{attn_arch}), which consists of a pre-trained action recognition ConvNet that extracts class-semantic rich features $R_{i,t} \in \{r_{1,t}, \dots , r_{C_{in},t}\}$ ($C_{in}$ represents the number of feature channels at each timepoint) at equidistantly distributed temporal points over the length of a video ($t \in \{1, \dots , T\}$). Action localization ConvNet (composed of sub-networks for encoder and localization) is trained with $R$ as input to generate action proposals (Fig.~\ref{attn_arch}). As the ConvNet is trained, the encoder sub-network learns to encode the class-semantic rich features $R$ to another temporally sequential feature $F_{i,t} \in \{f_{1,t}, \dots , f_{C_{out},t}\}$ more suitable for subsequent action localization tasks (e.g. generating the start and end temporal points of proposals and using confidence maps to group the start and end points \cite{bmn}), and the localization sub-network learns to use $F$ to perform these localization tasks.

In this work, we use standard action recognition, encoder, and localization sub-networks, and we only concentrate on the attention mechanism on $F$ (as shown in Fig.~\ref{attn_arch} by the pink dashed line). In Sec.~\ref{sec31}, we propose our novel attention mechanism that learns from the local temporal distribution of class-specific semantically rich features $R$ (or class semantics-based attention (CSA)) to estimate a usefulness or attention score for all feature channels and temporal points, which is then used to compute the attention-modified encoder feature $F_A$ as the new input to the localization sub-network. In Sec.~\ref{sec33}, we provide technical description of our attention mechanism. 

\subsection{Class semantics-based attention (CSA)} \label{sec31}
To help the localization sub-network attend more to feature channels and temporal locations that are more useful for localization tasks, there needs to be an attention mechanism that has the ability to adjust $F_{i,t}$ based on how useful or important it is to the action localization tasks. To mitigate the lack of input-adaptive channel/temporal attention mechanism in prior action localization networks, we propose an attention module that estimates the per-channel and per-timepoint attention weights independently and apply them independently on the localization encoder sub-network output $F$. We now discuss in detail about design of our attention module: its input, its joint learning framework, and the rationale behind this design choice, including the choice of input and the learning framework. 

\textbf{Input for the attention modules:} Based on prior works on attention, the more straightforward approach for applying attention on $F$ is to use $F$ itself, similar to self-attention frameworks such as the SE block \cite{squeeze}. The $F$ features are likely more potent for identifying background from foreground (having been trained for action localization), while the $R$ features are more potent for identifying which action class/classes the video belongs to (having been trained for action recognition). Assuming that our channel/temporal attention mechanisms are more dependent on the action class information-rich $R$ features rather than the foreground/background class information-rich $F$ features, $R$ is likely a better input choice for the attention modules, and we choose it as input to our attention modules. Comparing with self-attention modules that use $F$ as their input, we find that our CSA performs significantly better (Sec. 4.3), suggesting that such class semantics-based (or $R$ input) based attention mechanisms can be more potent. 

\textbf{Joint learning from class-semantics and temporal context:} In contrast to the task of image classification in the ablation study of \cite{squeeze} that had image as input, action localization tasks have video inputs that not only have class semantic information at different time points, but there is also variation of these class-semantics across the temporal axis. For example, `drinking coffee' action segment in a video can be preceded by `cooking' action and followed by `washing dishes' action. Thus, local temporal understanding of $R$ can provide indications on which class-semantics are more dominant at a time point. Also, since boundary detection is an important task for action localization network, temporal understanding of this variation of these class semantics could provide indications on which temporal positions are more important for boundary points.

\textbf{Computation of the attention weights:} The attention weights are computed by two separate attention modules, each of which have the class semantics rich $R$ feature as its input. Each of the attention modules (one estimating per-channel attention weight and the other estimating per-timepoint attention weight) learns from the local temporal variation of $R$ to compute the per-channel and per-timepoint attention weights. Then we apply our channel and temporal attention on $F$ (Fig.~\ref{attn_arch}) (or enable the localization sub-network to attend to more important feature channels and temporal locations to get $F_{A_C}$ and $F_{A_T}$, respectively (Fig.~\ref{attn_arch}). Finally, we aggregate $F_{A_C}$ and $F_{A_T}$ to get our final attention modified encoder output $F_A$ (Fig.~\ref{attn_arch}).
Next, we present the technical description of the attention module.

\subsection{Technical description of the CSA} \label{sec33}

We use a simple attention block consisting of a 1D convolution (with kernel size $k$) that operates succinctly to learn jointly from both the temporal and the channel axes of $R$. On the temporal axis, this 1D convolution has a receptive field of $k$, and on the channel axis, it has a receptive field of $C_{in}$ (or the entire length of feature channels). Thus, it captures information available from the global channel context as well as from the local temporal context simultaneously. Note that our CSA attention has two attention modules: one computes per-channel attention and the other computes per-time-point attention. To achieve this, we compute a channel attention vector and a temporal attention vector in parallel, assigning weights to specific channel-temporal input locations. Then, we perform late-fusion through a simple concatenation operation. We downsize the resultant output number of channels to that of class features $F$ using a simple projection layer to be subsequently processed by the localization sub-network. 

The input to our temporal attention block is the $R^{C_{in}\times T}$, where $C_{in}$ and $T$ correspond to number of input channels and temporal points respectively. Similarly, let $C_{out}$ denote number of output channels, and $F$ denotes the encoder feature which would be modified by our attention mechanism. Our temporal attention is computed as:
\begin{align}
\label{attn: Temporal}
\boldsymbol{y}_{T} &= {Avg}_{\{t=1,2,...,T\}}\left(\boldsymbol{f}_{{Tconv}_{C_{in} \times T}} ({R})\right) \\
\boldsymbol{A}_T &= sigmoid{\left(\boldsymbol{fc}_{T}(\boldsymbol{y}_T)\right)}\\
\boldsymbol{F}_{A_{T}} &= \left({A_T\mathbbm{1}_{1 \times C_{out}}}\right)^{T} \circ F 
\end{align}
where, $f_{{Tconv}_{C_{in} \times T}}$ is the core component of the temporal attention block, which has a 1D-convolutional layer of kernel size 3, a ReLU activation function. This block learns global channel context and local temporal context, and outputs a $T$ dimensional vector $y_T$. A fully connected layer(denoted by $\boldsymbol{fc}_{T}$)-sigmoid map on top of $\boldsymbol{y}_T$ yields $T$ dimensional attention vector $\boldsymbol{A}_T$. Finally, Hadamard operation with the encoder feature $F$ yields $F_{A_{T}}$, temporal attention modified encoder feature.

Similarly, we formulate our channel attention mechanism to obtain the channel attention modified encoder feature $F_{A_{C}} = {A_C\mathbbm{1}_{1 \times T}} \circ F$.

In the end, the attention modified encoder feature $F_{A}$ is computed using these two outputs. We compute this output by combining our channel attention modified encoder feature and temporal attention modified encoder feature by simply concatenating them, and mapping the number of channels back to $C_{out}$ using a convolution-ReLU block $f^{2 C_{out} \times C_{out}}_{TC}$. An alternative way to perform fusion is to add both the temporal and channel attended features together. However, this results in an averaging effect that diminishes the contribution from each of the attention modules.
Mathmatically, 
\begin{align}
\label{eq:1}
F_{A}^{C_{out} \times T} = f^{2 C_{out} \times C_{out}}_{TC}([F_{A_{T}};F_{A_{C}}])
\end{align}
\noindent where ``$;$" denotes the concatenation operator, and $F_{A}^{C_{out} \times T}$ denotes the final encoder output, modulated by temporal and channel attention mechanisms. The localization sub-network uses this output and performs subsequent steps required for localizing all the individual actions in the video. 

\section{Experiments and results}

\subsection{Experimental setup}
\noindent \textbf{Datasets:} To validate the utility of our CSA attention mechanism, we choose two popularly used action detection datasets - ActivityNet v1.3 \cite{activitynet} and THUMOS-14 \cite{thumos}. ActivityNet v1.3 is a larger-scale action localization benchmark (containing around 10K training and around 5K validation videos \cite{activitynet}) compared to the THUMOS-14 challenge dataset (containing 200 training and 213 validation videos \cite{thumos}). However, THUMOS-14 has more fine-grained action segment annotations overall (around 15 action instances per video) \cite{graph1,thumos} compared to ActivityNet v1.3 (around 1.65 action instances per video) \cite{activitynet}. 

\noindent \textbf{Evaluation metric for detection:} Mean Average Precision (mAP) is popularly used to quantitatively evaluate different approaches \cite{bmn,bsn,GTAD,GTAN,PBRNet,zhao2017temporal}, where the Average Precision (AP) is calculated on each action class respectively. We report the mAP at different Intersection over Union (IoU) thresholds and also report the overall mAP. For fair comparison between different networks, we use pre-trained and publicly available action recognition models (the CUHK classifier \cite{cuhk} for ActivityNet v1.3 \cite{activitynet} and UntrimmedNets \cite{wang2017untrimmednets} for THUMOS-14 \cite{thumos}), and we use the publicly available $R$ features (TSN features \cite{TSN} available in \cite{mmaction} for ActivityNet v1.3 \cite{activitynet} and TSN features \cite{TSN} available in \cite{GTAD} for THUMOS-14 \cite{thumos}). Finally, to evaluate proposal quality, we calculate Average Recall (AR) under different Average Number of proposals (AN) as AR$@$AN (AN varied from 1 to 100), and compute the area under the AR vs. AN curve (AUC).

\noindent \textbf{Implementation details:} We incorporate our CSA attention mechanism on three competitive and popularly used action localization networks, BMN \cite{bmn}, GTAD \cite{GTAD} and DBG \cite{DBG2020arXiv}. As is originally used in BMN \cite{bmn}, GTAD \cite{GTAD} and DBG \cite{DBG2020arXiv}, we use a temporal scale of $T=100$ for ActivityNet v1.3 and $T=256$ for Thumos14. Action localization model performances can also vary depending on the quality of the encoded features that are originally extracted from video sequences (i.e. the quality of the $R$ feature (Sec. 2.2)). Therefore, for fair comparison with other action detection benchmarks, we use the publicly available $R$ features (TSN features \cite{TSN} available in \cite{mmaction} for ActivityNet v1.3 \cite{activitynet} and TSN features \cite{TSN} available in \cite{GTAD} for THUMOS-14 \cite{thumos}). We use Adam optimizer with an initial learning rate of 0.001 with weight decay $1e^{-4}$ and train for 10 epochs on both the datasets using a step size of 7 for ActivityNet v1.3 and 3 for THUMOS-14. To compare utility of CSA against self-attention mechanisms, SE, CBAM and transformer~\cite{squeeze,cbam,transformer}, we implement SE, CBAM and transformer~\cite{squeeze,cbam,transformer} at the encoder output (self-attention i.e., attention applied on encoder output learns the attention weights based on the encoder output). To further evaluate whether our CSA benefits only from learning from processing the class-semantics rich encoder input features or whether it also benefits from the design of the CSA module, we experiment with replacing the CSA architecture with the SE architecture (FF-CSA in Table~\ref{attnablation}). 

\subsection{CSA's impact on action localization}

\paragraph{Performance improvement:} Incorporating CSA attention within baseline action localization ConvNets provides consistent performance gains on both the THUMOS-14 (3.8\% on GTAD baseline and 6.2\% on BMN baseline, Table \ref{thumosresults}) and the AcitivityNet v1.3 (0.52\% on GTAD baseline and 0.6\% on BMN baseline, Table \ref{activityresults}), and yields a new state-of-the-art (SotA) performance of 36.25\%mAP on the ActivityNet v1.3 dataset when used with BMN baseline~\cite{bmn} and publicly available TSP~\cite{tsp} encoded features. On the more challenging THUMOS-14 dataset that has higher number of action instances per video, the performance gains with CSA attention are significantly better than that on ActivityNet. This significant difference is perhaps due to a stronger necessity of using our CSA attention module (which helps the localization sub-network attend to important feature channels and temporal points) on the more difficult action detection tasks on videos that have higher number of action segments, boundaries and action classes. We also note that while the performance gains on THUMOS-14 across different tIoU are similar (e.g. 5.9\% improvement over BMN baseline at tIoU=0.3 and 6.2\% improvement over BMN baseline at tIoU=0.7), the performance actually drops on ActivityNet v1.3 for high tIoU=0.95. These action segments where the baseline BMN ConvNet performs better are mostly long duration action segments. Both of the above observations suggest that our CSA attention is particularly effective on challenging videos that either have higher number of action segments or have small duration action segments. In terms of proposal quality, DBG \cite{DBG2020arXiv} has the SotA AUC among prior approaches. We find that incorporating CSA within DBG improves AR@AN for all AN values (Table \ref{AUCvalues}) and improves overall AUC by 0.8\%.

\begin{table}[!htbp]
\begin{center}
\resizebox{\columnwidth}{!}{
\begin{tabular}{p{2.4cm}|p{0.6cm}p{0.6cm}p{0.6cm}p{0.6cm}p{0.6cm}p{0.6cm}}
\hline
 & & & \textbf{tIoU} & & & Avg  \\
\textbf{Method} & 0.3 & 0.4 & 0.5 & 0.6 & 0.7 & mAP \\
\hline\hline
R-C3D \cite{yan2017weakly} & 44.8 & 35.6 & 28.9 & - & - & -\\
SSN \cite{zhao2017temporal} & 51.9 & 41.0 & 29.8 & - & - & -\\
CBR \cite{gao2017cascaded} & 50.1 & 41.3 & 31.0 & 19.1 & 9.9 & 30.3\\
ETP \cite{qiu2018precise} & 48.2 & 42.4 & 34.2 & 23.4 & 13.9 & 32.4\\
BSN \cite{bsn} & 53.5 & 45.0 & 36.9 & 28.4 & 20.0 & 36.8\\
MGG \cite{MGG}& 53.9 & 46.8 & 37.4 & 29.5 & 21.3 & 37.8\\
BMN \cite{bmn} & 56.0 & 47.4 & 38.8 & 29.7 & 20.5 & 38.4\\
GTAD \cite{GTAD} & 54.5 & 47.6 & 40.2 & 30.8 & 23.4 & 39.3 \\
TAL-Net \cite{rethinking} & 53.2 & 48.5 & 42.8 & 33.8 & 20.8 & 39.8\\
CMS-RC3D\cite{bai2018contextual} & 54.7 & 48.2 & 40.0 & - & - & -\\
GTAN \cite{GTAN} & 57.8 & 47.2 & 38.8 & - & - & -\\
\hline
*GTAD-PGCN & 66.4 & 60.4 & 51.6 & 37.6 & 22.0 & 47.8 \\
\hline\hline
GTAD (Ours) & 54.5 & 48.2 & 39.9 & 30.4 & 21.2 & 38.8 \\
BMN (Ours) & 58.5 & 51.6 & 42.9 & 32.1 & 21.2 & 41.3 \\
GTAD-CSA & \textbf{58.4} & \textbf{52.8} & \textbf{44.0} & \textbf{33.6} & \textbf{24.2} & \textbf{42.6} \\
BMN-CSA & \textbf{64.4} & \textbf{58.0} & \textbf{49.2} & \textbf{38.2} & \textbf{27.8} & \textbf{47.5}\\
\hline
\end{tabular}
}
\end{center}
\caption{Quantitative comparison of the proposed CSA module with competing methods on THUMOS-14 action detection dataset in terms of mAP $@$tIoU. BMN+CSA beats all prior approaches by large margins. * denotes proposal-level post-processing methods that are complementary to other works including CSA.}
\label{thumosresults}
\end{table}
\begin{table}[!htbp]
\begin{center}
\begin{tabular}{l|c|c|c|c}
\hline
 & & \textbf{tIoU} & & Avg \\
\textbf{Method} & 0.5 & 0.75 & 0.95 & mAP \\
\hline\hline
R-C3D  \cite{yan2017weakly} & 26.80 & - & - & 12.70 \\
*CMS-RC3D\cite{bai2018contextual} & 32.92 & 18.36 & 1.13 & 18.46 \\
TCN \cite{dai2017temporal}  & 36.17 & 21.12 & 3.89 & - \\
*TAL-Net \cite{rethinking} & 38.23 & 18.30 & 1.30 & 20.22 \\
CDC \cite{cdc} & 43.83 & 25.88 & 0.21 & 22.77 \\
Xiong et al.\cite{xiong2017pursuit} & 39.12 & 23.48 & 5.49 & 23.98 \\
SSTAD \cite{sstad} & 44.39 & 29.65 & 7.09 & 29.17 \\
BSN \cite{bsn}& 46.45 & 29.96 & 8.02 & 30.03 \\
BMN \cite{bmn} & 50.07 & 34.78 & 8.29 & 33.85 \\
GTAD \cite{GTAD} & 50.36& 34.60& \textbf{9.02}& 34.09 \\
*GTAN \cite{GTAN} & 52.61 & 34.14 & 8.91 & 34.31 \\
DBG \cite{DBG2020arXiv} &  36.69&  26.61&  7.33& 25.67 \\
\hline
DBG (Ours) &  37.68&  27.39&  7.76& 26.43 \\
BMN (Ours) &  51.73&  35.74&  6.53& 34.83 \\
GTAD (Ours) & 51.45& 35.86& 8.86& 35.17\\
DBG-CSA & \textbf{40.08} & \textbf{29.45} &  \textbf{8.78}& \textbf{28.52} \\
BMN-CSA & \textbf{52.44} & \textbf{36.69} &  5.18& \textbf{35.43} \\
GTAD-CSA & \textbf{51.88} & \textbf{36.88} &  8.74& \textbf{35.69} \\
\hline
BMN-CSA-SE & \textbf{52.45} & \textbf{36.75} &  6.38 & \textbf{35.58} \\
BMN-CSA-Trans & 52.03 & 36.71 &  \textbf{9.23} & \textbf{35.69} \\
\hline
BMN (TSP \cite{tsp}) & \textbf{52.56} & \textbf{36.86} &  6.46& \textbf{35.89} \\
BMN-CSA (TSP) & \textbf{52.64} & \textbf{37.75} &  7.94& \textbf{36.25} \\
\hline

\hline
\end{tabular}
\end{center}
\caption{Quantitative comparison of the proposed CSA module with competing methods on ActivityNet v1.3 action detection dataset in terms of mAP $@$tIoU. BMN+CSA beats all prior two-stage approaches by notable margins and achieves SotA results, one-stage approaches are marked with * which are end-to-end trained from raw RGB frames.}

\label{activityresults}
\end{table}

\begin{table}[!htbp]
\begin{center}
\resizebox{\columnwidth}{!}{
\begin{tabular}{l|c|c|c|c|c}
\hline
Module & AR@1 & AR@5 & AR@10 & AR@100 & AUC (\%) \\
\hline
 BMN & 34.27 & 50.77& 57.27& 73.53& 65.98 \\
 DBG & 31.02 & 49.66& 57.45& 76.51& 68.19 \\
 BMN-CSA &\textbf{35.04} & \textbf{51.63} & \textbf{57.90} & \textbf{74.04} & \textbf{66.82} \\
 DBG-CSA &\textbf{32.46} & \textbf{50.48} & \textbf{58.27} & \textbf{77.12} & \textbf{68.89} \\
\hline
\end{tabular}
}
\end{center}
\caption{Comparison of AR@AN for baseline methods and methods with our CSA module on ActivityNet v1.3 dataset}
\label{AUCvalues}
\end{table}

\begin{table}[!htbp]
\begin{center}
\resizebox{\columnwidth}{!}{
\begin{tabular}{l|c|c|c|c|c|c}
\hline
Module & 0.3 & 0.4 & 0.5 & 0.6 & 0.7 & mAP \\
\hline
 Trans & 55.1 & 48.8& 41.3& 30.5& 19.7& 39.1 \\
  CBAM & 56.8&50.3&42.3&31.9&22.3 &40.7\\
 SE & 62.4 & 54.8& 46.1& 34.2& 23.6& 44.2 \\
 Ours &\textbf{64.4} & \textbf{58.0} & \textbf{49.2} & \textbf{38.2} & \textbf{27.8} & \textbf{47.5} \\
\hline
\end{tabular}
}
\end{center}
\caption{Comparison of other attention modules with our CSA module on THUMOS-14}
\label{selfattnablation}
\end{table}

\begin{figure}
\begin{center}
\includegraphics[height=5cm, width=6cm]{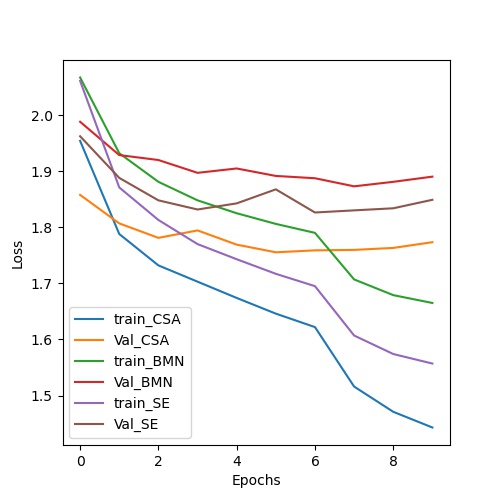}\\
\end{center}
   \caption{Comparison of training and validation loss curves of CSA with SE attention and baseline BMN on ActivityNet v1.3.}
\label{LossCurve}
\end{figure}

\paragraph{CSA accelerates training:} Incorporating CSA attention on baseline BMN and GTAD speeds up convergence of the networks with faster reduction in both training and validation losses. Fig.~\ref{LossCurve} shows the comparison of loss curves for baseline and SE BMN models with CSA. We find that CSA has a steeper loss curve for both training and validation as compared to SE attention suggesting that CSA helps the network to learn better. To test whether the mAP performance improvement from incorporating CSA is simply from accelerated training, we train baseline BMN for higher number of epochs (20 instead of original 10 epochs). We find no improvement in baseline BMN performance, suggesting that our CSA attention module not only accelerates training but also leads to better optimized networks for action detection.

\paragraph{Computational considerations:} The added computation cost with CSA attention (around 1ms/video) is similar to that of the efficient SE~\cite{squeeze} mechanism. This is because the architecture of CSA is mostly similar to that of SE except that it includes an extra lightweight 1D convolution that learns jointly from both class-semantics and temporal context. Note that CSA provides considerably more mAP gain compared to SE (Table \ref{selfattnablation}).

\begin{figure*}
  \begin{subfigure}{\linewidth}
  \includegraphics[width=\linewidth]{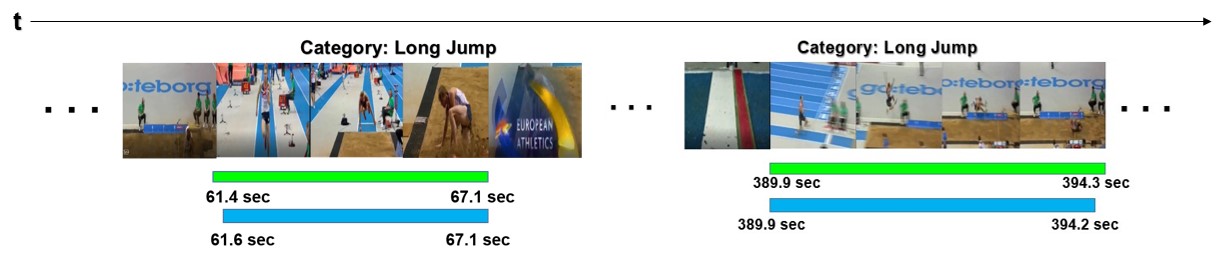}\\
  \includegraphics[width=\linewidth]{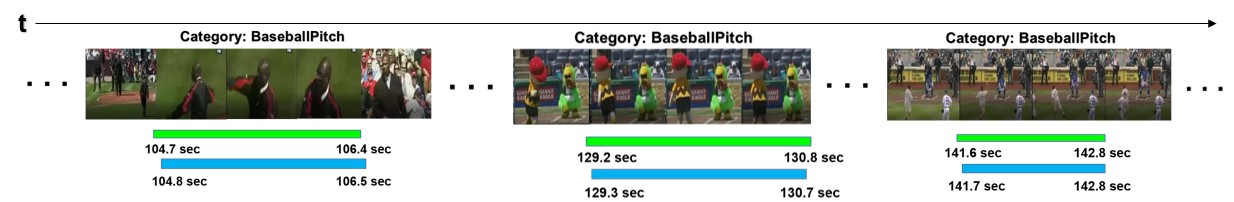}
  \end{subfigure}\par\medskip
  \caption{Qualitative results from THUMOS-14 dataset. Green segments show groundtruth and blue segments show predicted segments using BMN-CSA model. Our model is able to accurately detect diverse length sport domain actions.}
  \label{fig:qualthumos}
\end{figure*}

\subsection{Ablation studies}
\paragraph{Comparison with self-attention mechanisms:} CSA is different from self-attention mechanisms in two ways: (1) CSA estimates the attention weights based on class-semantics rich features at the encoder input and applies it at the encoder output whereas self-attention mechanisms applied at the encoder output estimates the attention weights based on localization-information rich features at the encoder output itself, (2) CSA learns jointly from both class-semantics and temporal context facilitated by a simple 1D convolution. Our CSA attention mechanism outperforms three popular self-attention mechanisms we experimented with, namely the transformer (Trans)~\cite{transformer}, CBAM~\cite{cbam} and SE \cite{squeeze} attention mechanisms when applied at the encoder output of the BMN (Table~\ref{selfattnablation}). CSA outperforms all of these attention mechanisms. Interestingly, we find the training loss with including Transformer in BMN to be lower than BMN without the attention module, yet the validation mAP with Transformer is lower showing signs of over-fitting. This is not the case for our CSA attention or SE/CBAM, all of which provide mAP gains as well. To find whether self-attention mechanisms and our CSA can be complementary, we incorporate SE after all the 1D convolution layers in the BMN encoder, except at the encoder output where we incorporate the CSA attention. As shown in Table~\ref{activityresults} we find that this provides additional gain of \textbf{0.15\%} mAP. We also conduct the same experiment with Transformer instead of SE and find an additional gain of \textbf{0.26\%} mAP. This complementary behavior between the two attention mechanisms (i.e. CSA with self-attention) is perhaps due to the difference in the inputs - SE/Transformer learns from the localization-rich features within the encoder sub-network, whereas CSA learns from the class-semantics rich features at the encoder input $R$. 

\begin{table}[!htbp]
\begin{center}
\resizebox{\columnwidth}{!}{
\begin{tabular}{l|c|c|c|c|c|c}
\hline
Module & 0.3 & 0.4 & 0.5 & 0.6 & 0.7 & mAP \\
\hline
  FF-CSA& 60.9 & 54.1 & 45.8 & 35.5 & 24.1 & 44.1\\
 G-CSA & 61.7 & 54.9& 46.2& 36.2& 24.7& 44.8 \\
 CSA (1-Conv) & \textbf{64.1} & \textbf{56.3}& \textbf{48.2}& \textbf{38.1}& \textbf{26.6}& \textbf{46.7} \\
 LG-CSA & 59.8 & 53.2& 44.4& 34.5& 24.4& 43.3 \\ 
\hline
\end{tabular}
}
\end{center}
\caption{Ablation study of using different variations of CSA module on THUMOS-14. Note: CSA 1-Conv is used for fair comparison with other baselines having single convolutional layer.}
\label{attnablation}
\end{table}

\begin{table}[!ht]
\begin{center}
\resizebox{\columnwidth}{!}{
\begin{tabular}{l|c|c|c|c|c|c|c}
\hline
Channel & Temporal & 0.3 & 0.4 & 0.5 & 0.6 & 0.7 & mAP \\
\hline
\checkmark & & 62.1 & 55.8& 46.4& 36.5& 24.7& 45.1 \\
&\checkmark & 62.5 & 56.2& 46.5& 36.9& 26.1& 45.7 \\
 \checkmark  & \checkmark & \textbf{64.1} & \textbf{56.3}& \textbf{48.2}& \textbf{38.1}& \textbf{26.6}& \textbf{46.7} \\
\hline
\end{tabular}
}
\end{center}
\caption{Ablation study of different components of CSA (with one convolution) on THUMOS-14 dataset. }
\label{componentAttn}
\end{table}

\begin{table}[!htbp]
\begin{center}
\begin{tabular}{l|c|c|c|c|c|c}
\hline
Module & 0.3 & 0.4 & 0.5 & 0.6 & 0.7 & mAP \\
\hline
 k=1 & 62.8 & 55.8& 47.2& 36.4& 24.9& 45.4 \\
  k=3& \textbf{64.1} & \textbf{56.3}& \textbf{48.2}& \textbf{38.1}& \textbf{26.6}& \textbf{46.7} \\
 k=5 & 57.9 & 49.5& 40.4& 30.8& 20.5& 39.8 \\
 \hline
 1-Conv & 64.1 & 56.3& 48.2& 38.1& 26.6& 46.7 \\
 2-Conv & \textbf{64.4} & \textbf{58.0} & \textbf{49.2} & \textbf{38.2} & \textbf{27.8} & \textbf{47.5} \\
 3-Conv & 61.8 & 54.7& 45.3& 35.5& 23.9& 44.2 \\
\hline
\end{tabular}
\end{center}
\caption{Effect of different kernel size and number of convolutional blocks inside CSA module on THUMOS-14}
\label{kernelablation}
\end{table}
\paragraph{Alternate design choices for CSA} \label{alternateAC}
We experiment with different design choices for computing the attention weights via processing of the input features $R$ with different variations to the 1D convolution design choice. These variants are: (1) Direct (FF) - not using the 1D convolution at all, (2) global context learning (G) - using a transformer encoder \cite{transformer} that has global receptive field instead of the convolution block. (3) Local context learning (L) - We use 1D convolution to locally process the temporal scales of the input features for computing the attention weights. (4) Local-Global Context learning (LG) - To see if local and global information complement each other, we apply local 1D convolutional operation followed by a global transformer operation. Tables \ref{attnablation}, \ref{componentAttn} show the results of using different variants and we find that local information (i.e., local temporal features) is key for computing the attention weights as compared to others. While global information can perhaps help in providing better attention weights, we conjecture that since learning such global information requires modules with higher number of learnable parameters (e.g. transformer encoder requires more parameters than a simple 1D convolution), it becomes difficult to train the attention sub-network. On that note, we also experiment with the hyper-parameters within the attention module (Table~\ref{kernelablation}) by increasing the number of sequential 1D convolution - we find supporting evidence in here as well in that increasing the number of sequential 1D convolution from one to two improves the results whereas from two to three decreases the performance significantly. Similarly, increasing the kernel size from 3 to 5, which increases the number of learnable parameters, leads to inferior performance, suggesting that larger CSA modules with higher number of learnable parameters could be difficult to train.
\begin{table}[!htbp]
\begin{center}
\resizebox{\columnwidth}{!}{
\begin{tabular}{l|c|c|c|c|c|c}
\hline
Module & 0.3 & 0.4 & 0.5 & 0.6 & 0.7 & mAP \\
\hline
  Start-CSA& 60.9 & 53.6 & 44.9 & 34.3 & 23.3 & 43.4\\
 Middle-CSA & 63.8 & 57.1& 47.7& 36.9& 25.3& 46.2 \\
 CSA (2-Conv) & \textbf{64.4} & \textbf{58.0} & \textbf{49.2} & \textbf{38.2} & \textbf{27.8} & \textbf{47.5}\\
\hline
\end{tabular}
}
\end{center}
\caption{Ablation study of applying the CSA module at different locations of the encoder module on THUMOS-14 dataset. Start-CSA applies attention at the encoder input. Middle-CSA applies attention at the middle of the encoder.}
\label{locationablation}
\end{table}
\paragraph{Location of application of CSA attention}
We conduct an ablation to study the effect of applying the CSA attention module at different locations within the encoder module. We identify three cases - (1) At the start (2) At the middle and (3) At the output of the encoder. Table \ref{locationablation} shows the results of applying the CSA attention module at the three locations. We find that the attention applied at the end of the encoder (input to the localization sub-network) outperforms other results. CSA attention applied to the input features gives significantly lower (3.9\% drop) results. This shows that applying attention at $F$ optimally modulates the features such that it helps to localize the start/end times and their corresponding scores more precisely, while not interfering with the encoder sub-network that has a separate objective of learning a better feature encoding process for the input $R$. 
\section{Discussion}
We find CSA to consistently provide performance gains when incorporated with publicly available and readily reproducible baseline networks~\cite{bmn,DBG2020arXiv,GTAD} on popular action localization datasets (THUMOS-14 and ActivityNet v1.3) with different sets of publicly available encoded features (e.g. TSN features~\cite{TSN}, TSP features~\cite{tsp}). We also find that CSA is better than self-attention mechanisms, and in addition provides complementary gains when used with self-attention mechanisms. Our BMN-CSA with TSP features currently have the SotA in ActivityNet v1.3 and was a core technical contribution behind achieving a top position in the 2021 ActivityNet temporal localization challenge. Further, Fig.~\ref{fig:qualthumos} shows qualitative results of the SotA BMN-CSA model on THUMOS-14 dataset.~From the figure, it can be observed that our model is accurately able to detect action segments of two different categories of different action lengths. Despite these promising results on consistent performance gains and current SotA, our CSA evaluation is currently limited in not being tested with other competitive baseline architectures that have been shown to have better performance than BMN~\cite{bmn}, e.g. GTAD-PGCN~\cite{GTAD}. We envision that incorporating CSA in such improved network architectures will yield better SotA performances.
\section{Conclusion}
In this paper, we proposed a novel attention mechanism, CSA, for action detection networks that utilizes action-class semantics to compute attention weights. CSA mechanism is effective in applying attention along both the channel and the temporal axes of the encoded feature of an action localization network. We demonstrate that CSA is generic and is easily integrated with prior action detection approaches. Incorporating CSA provides consistent performance gains on action detection tasks, and incorporating it in prior SotA benchmarks yields new SotA results - e.g. 36.25\% mAP on the ActivityNet v1.3. 

{\small
\bibliographystyle{ieee_fullname}
\bibliography{egbib}
}

\end{document}